%% file: Arxiv.tex
\documentclass[10pt,twocolumn,letterpaper]{article}

\usepackage[pagenumbers]{cvpr} 

\usepackage{graphicx}
\usepackage{amsmath}

\usepackage{amssymb}
\usepackage{booktabs}
\usepackage{subcaption}
\usepackage{multirow}
\usepackage[symbol]{footmisc}
\usepackage[accsupp]{axessibility}
\usepackage{bm}
\usepackage{float}
\usepackage{graphics}
\usepackage{multirow}
\usepackage{fixmath}

\usepackage{subcaption}
\usepackage{graphicx}
\usepackage{algpseudocode}
\usepackage{pifont}
\usepackage{mathtools}
\usepackage{xfrac}
\usepackage[dvipsnames]{xcolor}
\usepackage{arydshln}

\usepackage[pagebackref,breaklinks,colorlinks]{hyperref}
\usepackage{bookmark}

\usepackage[capitalize]{cleveref}
\crefname{section}{Sec.}{Secs.}
\Crefname{section}{Section}{Sections}
\Crefname{table}{Table}{Tables}
\crefname{table}{Tab.}{Tabs.}

\newcommand{\Eref}[1]{Eq.~(\ref{#1})}

\DeclareMathOperator*{\argmin}{arg\,min}

\newcommand{\minisection}[1]{\vspace{2mm}\noindent{\textbf{#1}}}

\def\cvprPaperID{11434} 
\def\confName{CVPR}
\def\confYear{2023}

\begin{document}
\input{definitions}

\title{HyperCUT: Video Sequence from a Single Blurry Image \\ using Unsupervised Ordering}

\author{
Bang-Dang Pham$^{1,*}$ \quad Phong Tran$^{1,2,*}$ \quad Anh Tran$^{1}$ \quad Cuong Pham$^{1,3}$ \quad Rang Nguyen$^{1}$\quad Minh Hoai$^{1,4}$\\ 
\small{\textsuperscript{1}VinAI Research, Vietnam \quad \textsuperscript{2}MBZUAI, UAE \quad \textsuperscript{3}Posts \& Telecommunications Inst. of Tech., Vietnam
\quad \textsuperscript{4}Stony Brook University, USA}\\
\texttt{\scriptsize \{v.dangpb1, v.anhtt152,  v.hoainm\}@vinai.io} \quad 
\texttt{\scriptsize cuongpv@ptit.edu.vn   }
\texttt{\scriptsize  the.tran@mbzuai.ac.ae} \\
\small{\textsuperscript{*}Equal contribution}
}
\maketitle
\begin{abstract}
   We consider the challenging task of training models for image-to-video deblurring, which aims to recover a sequence of sharp images corresponding to a given blurry image input. A critical issue disturbing the training of an image-to-video model is the ambiguity of the frame ordering since both the forward and backward sequences are plausible solutions. This paper proposes an effective self-supervised ordering scheme that allows training high-quality image-to-video deblurring models. Unlike previous methods that rely on order-invariant losses, we assign an explicit order for each video sequence, thus avoiding the order-ambiguity issue. Specifically, we map each video sequence to a vector in a latent high-dimensional space so that there exists a hyperplane such that for every video sequence, the vectors extracted from it and its reversed sequence are on different sides of the hyperplane. The side of the vectors will be used to define the order of the corresponding sequence. Last but not least, we propose a real-image dataset for the image-to-video deblurring problem that covers a variety of popular domains, including face, hand, and street. Extensive experimental results confirm the effectiveness of our method. Code and data are available at \url{https://github.com/VinAIResearch/HyperCUT.git}
\end{abstract}

\section{Introduction}
Motion blur artifacts occur when the camera's shutter speed is slower than the object's motion. This can be studied by considering the image capturing process, in which  the camera shutter is opened to allow light to pass to the camera sensor. This process can be formulated as: 
\begin{align}
    y = g\left(\frac{1}{\tau} \int_{0}^{\tau} x(t) dt\right) \approx g\left(\frac{1}{N+1}\sum_{k=0}^{N} x_k\right),
    \label{blur_formula}
\end{align}
where $y$ is the resulting image, $x(t)$ is the signal captured by the sensor at time $t$, $g$ is the camera response function, and $\tau$ is the camera exposure time. For simplicity, we omit the camera response function in the notation. 
The image $y$ can also be approximated by averaging $N{+}1$ uniform samples of the signal $x$, denoted as $x_k$ with $k = \overline{0,N}$. 
For long exposure duration or rapid movement, these samples can be notably different, causing motion blur artifacts.

Image deblurring seeks to remove the blur artifacts to improve the quality of the captured image. This task has many practical applications, and it has been extensively studied in the computer vision literature. However, existing methods often formulate the deblurring task as an image-to-image mapping problem, where only one sharp image is sought for a given blurry input image, even though a blurry image corresponds to a sequence of sharp images. The image-to-image approach can improve the aesthetic look of a blurry image, but it is insufficient for many applications, especially the applications that require recovering the motion of objects, e.g., for iris or finger tracking. In this paper, we tackle the another important task of image-to-video deblurring, which we will refer to as  \textbf{$blur2vid$}. 

\begin{figure}[t]
    \small
    \begin{center}
        \includegraphics[width=.475\textwidth]{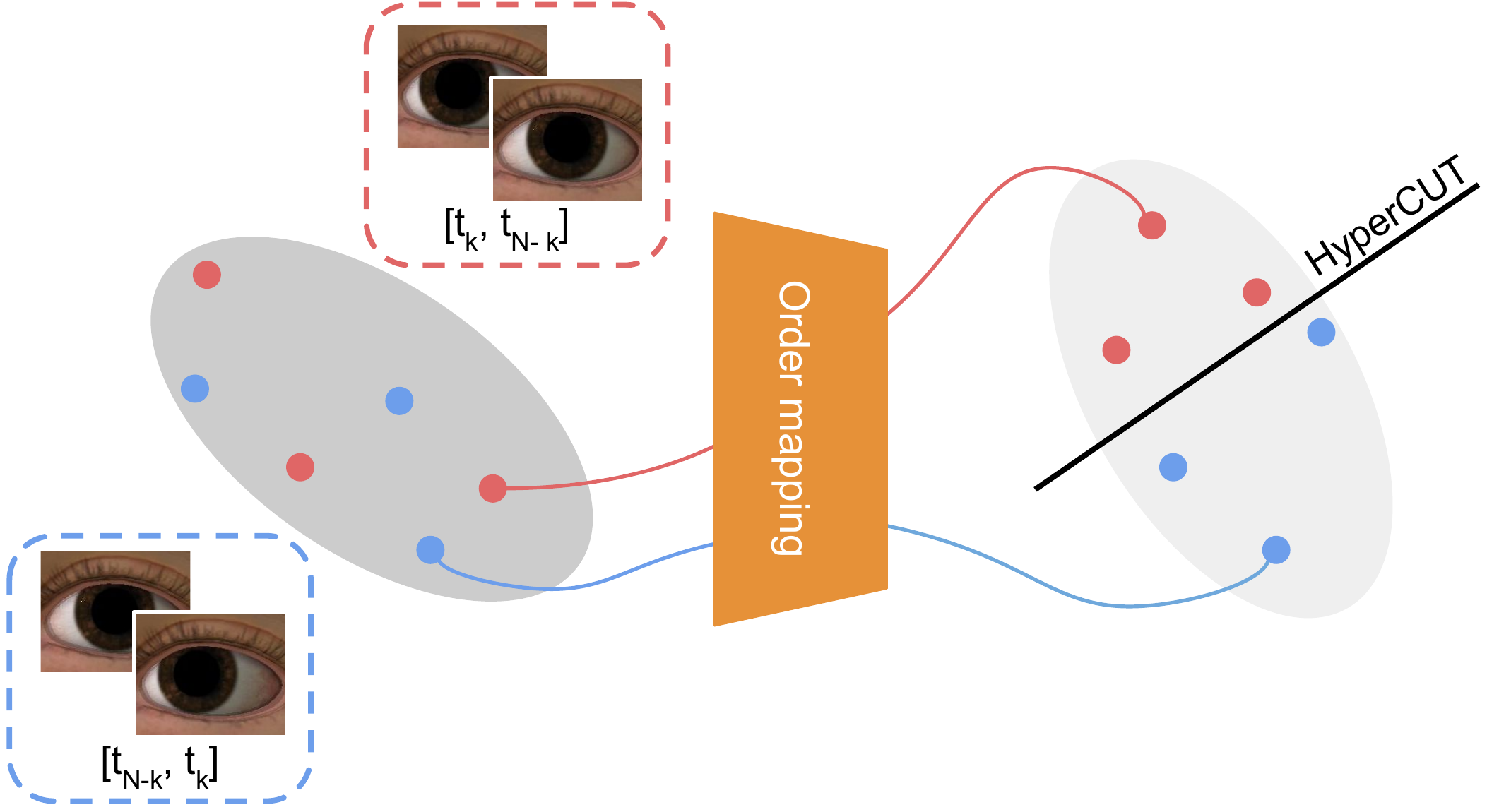}
    \end{center}
    \vskip -0.2in
    \caption{We tackle the order ambiguity issue by forcing the frame sequence to follow a pre-defined order. To find such an order, we map the frame sequences into a high dimensional space so that they are separable. The side (left or right of the hyperplane) is used to define the order of the frame sequence.}
    \label{fig:teaser}
    \vskip -0.12in
\end{figure}

Image-to-video deblurring, however, is a non trivial task that requires learning a set of deblurring mapping functions $\{f_k\}$ so that $f_k(y) \approx x_k$. A naive approach is to minimize the squared difference between the predicted sharp image and the ground truth target, i.e.,
\begin{align}
f_k = \myargmin{f} \bbE_{x, y} ||f(y) - x_k||_2^2.	\label{eqn:eq2}
\end{align}

However, this approach suffers from the order-ambiguity issue \cite{jin2018learning}. Considering \Eref{blur_formula}, the same blurry image $y$ is formed regardless of the order of the corresponding sampled sharp frames. For example, both $\{x_0,  ..., x_{N}\}$ and the reversed sequence are valid solutions. Thus, $x_k$, $x_{N-k}$, and possibly other $x_h$'s are valid `ground truth' target for $f_k(y)$. Thus, optimizing \Eref{eqn:eq2} will lead to a solution where $f_k$ is mapped to the average of $x_k$ and $x_{N-k}$. This issue has also been observed in the work of \cite{mathieu2015deep} for future video frame prediction. This also explains why most existing deblurring methods cannot be directly used to recover any frame other than the middle one. To tackle this issue, Jin 
\etal~\cite{jin2018learning} introduced the order-invariant loss, \Eref{ordering_inv_loss}, which computed the total loss on frames at symmetric indexes (i.e., $k$ and $N-k$). However, this loss does not fully resolve the issue of having multiple solutions, as will be demonstrated in \cref{sec:toy_example}. 

This paper proposes a new scheme to solve the order ambiguity issue. Unlike the order-invariant loss \cite{jin2018learning} or motion guidance \cite{zhong2022animation}, we solve this problem directly by explicitly assigning which frame sequence is backward or forward. In other words, each sequence is assigned an order label $0$ or $1$ so that its label is opposite to the label of its reverse. Then the ambiguity issue can be tackled by forcing the model to learn to generate videos with the order label  ``0''. We introduce {\bf HyperCUT} as illustrated in \cref{fig:teaser} to find such an order. Specifically, we find a mapping $\mathcal{H}$ that maps all frame sequences into a high-dimensional space so that all pairs of vectors representing two temporal symmetric sequences are separable by a hyperplane. We dub this hyperplane HyperCUT. Each frame sequence's order label is defined as the side of its corresponding vector w.r.t. the hyperplane. We find the mapping $\mathcal{H}$ by representing it as a neural network and training it in an unsupervised manner using a contrastive loss.

Previously, there existed no real \textit{$blur2vid$} dataset, so another contribution of this paper is the introduction of a new dataset called Real $blur2vid$ ({\bf RB2V}).  RB2V was captured by a beam splitter system, similar to \cite{rim2020real,zhong2021towards,zhong2020efficient}. It consists of three subsets for three categories: street, face, and hand. We will use the last two to demonstrate the potential applications of the $blur2vid$ task in motion tracking. 

In short, our contributions are summarized as follow:
\begin{itemize} \denselist 
    \item We introduce HyperCUT which is used to solve the order ambiguity issue for the task of extracting a sharp video sequence from a blurry image. 
    \item We build a new dataset for the task, covering three categories: street, face, and hand. This is the first real and large-scale dataset for image-to-video deblurring.
    \item We demonstrate two potential real-world applications of image-to-video deblurring.
\end{itemize}

\section{Related Work}
\subsection{Image deblurring}
Image deblurring is a classical task in low-level computer vision. In the past, the blur kernel was assumed to be linear and uniform, and the blur model can be formulated as: $y = x * k + \eta$, 
where $k$ is the blur kernel, $x$ is the sharp image, $*$ denotes convolution operator, $\eta$ is the white noise,  and $y$ is the corresponding blurry one. The main approach was to find a good prior for either the sharp images \cite{pan2017deblurring,krishnan2009fast,chan1998total,xu2013unnatural,krishnan2011blind} or the blur kernel \cite{liu2014blind} space. However, the complexity of the optimization involved in these methods, along with their reliance on linear and uniform assumptions, renders them unsuitable for generalizing to real-world blur scenarios.

Thanks to the advance of deep neural networks in the past few years, the community has witnessed a significant leap in the deblurring field. Deep learning-based models do not make any explicit assumption on the blur operator nor on the sharp image space. Instead, they can learn to deblur using large-scale datasets. Zamir \etal \cite{zamir2021multi} proposed a multi-stage architecture, where contextual information was learned in the earlier stages. In contrast, the whole input image was processed without any downsample operator to extract fine spatial details in the last stage. Tao \etal~\cite{tao2018scale} employed a multi-scale recurrent network that deblurred the input image in a multi-scale and recurrent manner. Kupyn \etal \cite{kupyn2018deblurgan,kupyn2019deblurgan} introduced generative adversarial networks \cite{goodfellow2020generative} for the deblurring task to make the deblurred image more realistic. However, the performance of deep deblurring models degrades significantly when the blur operator does not appear in the training set~\cite{tran2021explore}.

\subsection{Recovery of multiple sharp frames \label{sec:toy_example}}
Jin \etal~\cite{jin2018learning} were the first to introduce a model that took a blurry input $y$ and produced multiple sharp frames $x_0, \ldots, x_6$. They trained seven networks $f_0, \ldots, f_6$
, each corresponded to a sharp frame output. Their method was also the first to point out the order-ambiguity issue: finding a set of sharp frames given a single blurry input is an ill-posed problem since the generation of $y$ is independent of the order of the sharp frames. \cite{jin2018learning} addressed this by introducing the order-invariant loss:
\vspace{-2mm}
\begin{multline}
    \mathcal{L}_{OI} = \sum_{k=0}^{2} \left( \left| \|f_k(y) - f_{6 - k}(y)\| - \|x_k - x_{6 - k}\| \right| \right.\\
             + \left. \left| \| f_k(y) + f_{6 - k}(y)\| - \|x_k + x_{6 - k}\| \right| \right).
    \label{ordering_inv_loss}
\end{multline}

\begin{figure}[t]
    \small
    \begin{center}
    \includegraphics[scale=0.087]{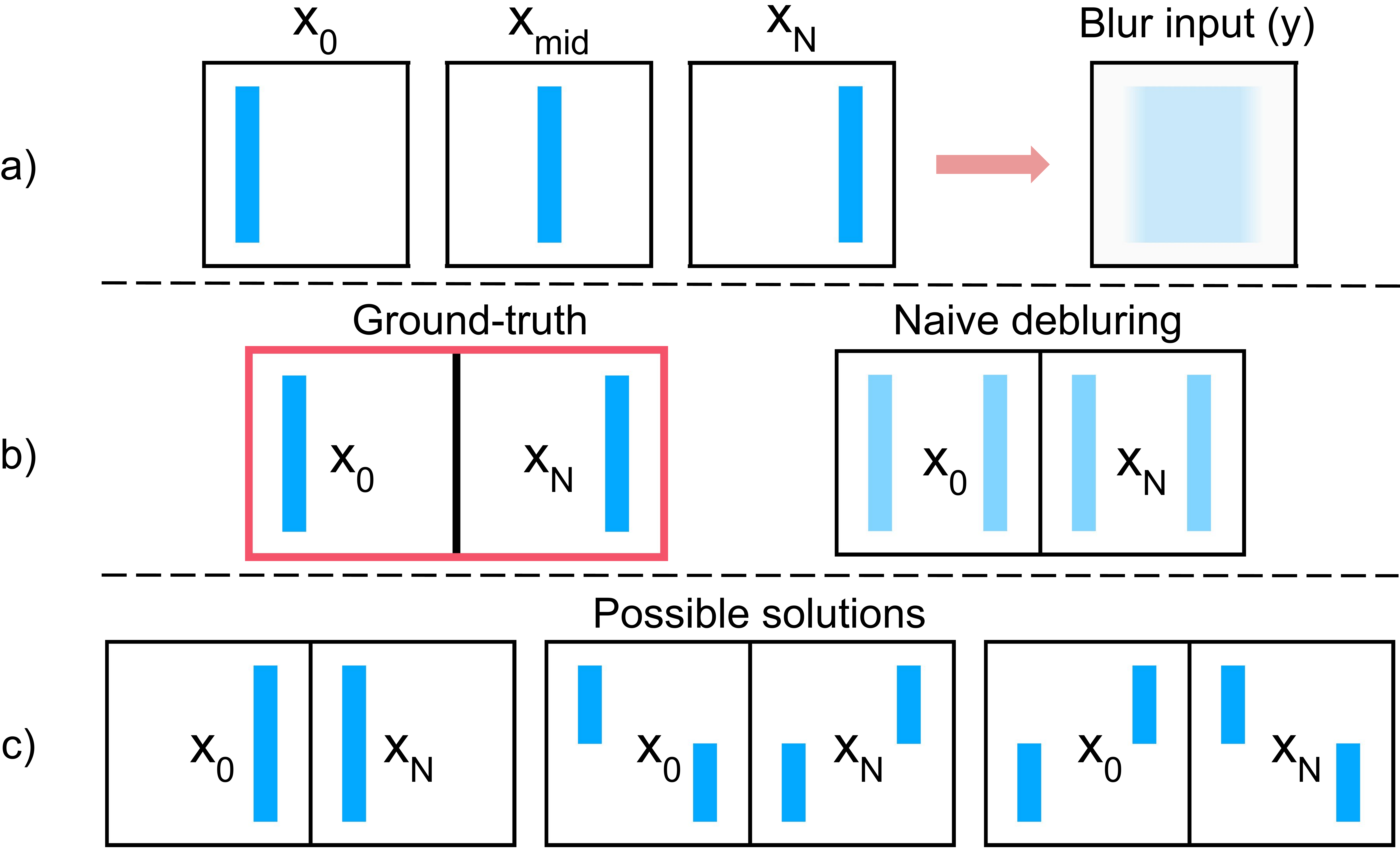}\\
    \end{center}
    \vskip -0.2in
\caption{\textbf{Toy example.} Row (a) depicts the formation of a blurry image $y$ from a sharp  sequence. We consider the task of recovering border frames ($x_0$, $x_N$) from $y$. The ground-truth label is provided in the first column of Row (b). Due to the order-ambiguity issue, normal regression networks often return the blurry result as in the second column of Row (b). Order-invariant loss \cite{jin2018learning} accepts both the correct solution and three other ones in Row (c). Our proposed method only returns the correct solution (red-box).}
    \label{fig:multiple_solution}
    \vspace{-5mm}
\end{figure}

However, this loss does not fully address the issue as illustrated in \cref{fig:multiple_solution}. Row (a) shows the formation of the blurry image $y$ from a sequence of sharp images. Consider the sub-task of recovering the end frames ($x_0$, $x_N$) from the blurry one. The ground truth solution for this task is shown in the first column of Row (b). Due to the order-ambiguity issue, normal regression networks often return the blurry solution as in the second column of Row (b), in which each predicted frame is an average of the ground-truth pair. By applying the order-invariant loss~\cite{jin2018learning}, one can get sharp prediction outcomes. However, besides the correct frame pair, it also accepts three other solutions in Row (c). The first solution is just different from the ground truth by the frame order, while the other two are obviously wrong.

\setlength{\tabcolsep}{2pt}

Purohit \etal\cite{purohit2019bringing} proposed a recurrent architecture that could be extended to generate any number of sharp frames without increasing the number of parameters. They trained a pair of recurrent video encoder and decoder to reconstruct a set of $N$ continuous sharp frames. The encoder was then replaced by a blurred image encoder to form a network that could generate $N$ sharp frames from a single blurry image. All these methods rely on the order-invariant loss \cite{jin2018learning} to avoid the order-ambiguity issue. However, this loss does not fully address the issue. The models proposed by \cite{pan2019bringing,xu2021motion} do not suffer from order ambiguity, but they need additional data from an event camera.

Recently, Zhong \etal~\cite{zhong2022animation} proposed a different approach to solving the order ambiguity issue. Instead of focusing on the training loss, they converted the ill-posed $blur2vid$ task into a nearly deterministic one-to-one mapping problem by using motion guidance as an additional input. This motion guidance input was generated from the blurry image by a conditional Variation Autoencoder such that it was unique for each solution. However, the model largely depended on the quality of the motion guidance and consequently failed when the human-annotated data was not available or the estimated optical flow was inaccurate. In addition, the motion guidance was built upon handcrafted heuristics and might not hold for every case, especially for complicated motion.

\subsection{Real blur datasets}
To train deep deblurring models, many large-scale sharp-blur pair datasets have been proposed. Tao \etal~\cite{tao2018scale} introduced the GOPRO dataset, which consists of more than 1000 pairs of sharp images captured by a high-speed GOPRO 4 Hero Black and their corresponding synthetic blurry images. They generated blurry images by mimicking the blur generation process as described in \Eref{blur_formula}. Nah \etal~\cite{nah2019ntire} proposed the REDS dataset with a similar synthesis method, but with more pairs, higher quality, and a different camera response function choice. \cite{shen2019human} proposed a human-aware deblurring dataset that focused on human movements. Tran \etal~\cite{tran2021explore} used a blur encoder to transfer blur operator from existing datasets to another sharp frame set. Zhong \etal~\cite{zhong2022animation} proposed B-Aist$++$ dataset, which was synthesized from dancing videos \cite{li2021ai} to simulate complex human body movements.

Since deep deblurring models are highly overfitted to the blur operator used in the training dataset \cite{tran2021explore}, real-image deblurring datasets are critical. Therefore, many have been introduced over the past few years \cite{rim2020real,zhong2021towards,zhong2020efficient}. These datasets were captured by a system that consists of high and low shutter speed cameras. Two cameras were placed on two sides of a beam splitter to capture the same scene. Existing datasets used for image/video deblurring are not sufficient to train \textit{$blur2vid$} models. Jin \etal \cite{jin2018learning} built a synthetic dataset by using seven consecutive frames and their average as ground-truth and input, respectively. To the best of our knowledge, there was no real-image deblurring dataset for the $blur2vid$ task.

\begin{figure*}[t]
    \centering
    \includegraphics[width=.85\textwidth]{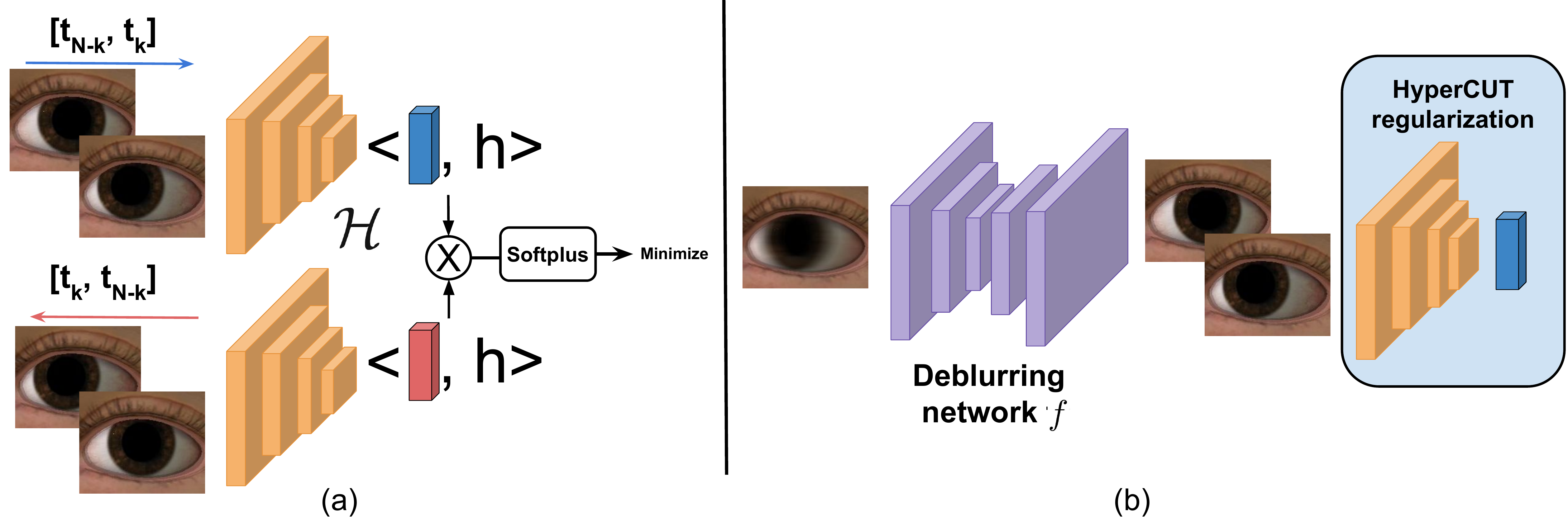}
    \vspace{-2mm}
    \caption{The overall architecture of our framework. (a) Given a frame sequence, we optimize the function $\mathcal{H}$ by forcing the sides of the vectors representing it and its reverse to be different, according to a fixed hyperplane $h$. (b) Having $\mathcal{H}$, we can use it as a regularization to make the deblurring network only predict frame sequences of the same side of $h$, thus solving the order-ambiguity issue.}
    \label{fig:overall_architecture}
    \vspace{-3mm}
\end{figure*}

\vspace{-1mm}
\section{Methodology}
\vspace{-1mm}

This section describes the proposed method. We assume there is training data of the form $\{(y^i, x_0^i, \ldots, x_N^i) \}_{i=1}^{M} $, where $M$ is the number of training samples, and each training sample consists of a blurry image $y^i$ and $N+1$ sharp images. Our goal is to train neural networks that can recover all the sharp images from the blurry one. 

\subsection{HyperCUT order}\label{sec:hypercut}
\vspace{-0.5mm}
One approach for the \textit{$blur2vid$} task is to pose it as multiple image-to-image deblurring tasks and train a separate network for each task. It means that for each target frame index $k \in [0..N]$, we train a network $f_k$ to predict $x^i_k$ from $y^i$ by optimizing:  

\vspace{-5mm}
\begin{align}
    \mathcal{L}_{rec} = \frac{1}{M}\sum_{i=1}^{M}\|f_k(y^i) - x_k^i\|. \label{eqn:Lrec}
\end{align}

Unfortunately, this naive approach fails to produce a sharp output. Empirically, we observe that this approach tends to output an image that is close to $\frac{(x_k + x_{N-k})}{2}$. Conceptually, it is known that the output of $f_k$ will converge to the average of all sampled targets used for training, which include both $x_k$ and $x_{N-k}$ due to the order-ambiguity issue. 

Let $h$ be a fixed hyperplane in a high-dimensional space. We want to find a mapping $\mathcal{H}: \bbR^{2 \times H \times W \times C} \rightarrow \bbR^d$ such that $\mathcal{H}\left([x^i_k, x^i_{N - k}]\right)$ and $\mathcal{H}\left([x^i_{N - k}, x^i_k]\right)$ are on different sides of $h$. In other words:
\begin{align}
    \left< \mathcal{H}\left([x^i_k, x^i_{N - k}]\right), h\right>   \left<\mathcal{H}\left([x^i_{N - k}, x^i_k]\right), h\right> < 0,
\label{eq:dot_prod}
\end{align}
To find the mapping $\mathcal{H}$, we represent it by a neural network as shown in \cref{fig:overall_architecture}. The objective function is to minimize the left hand side of \Eref{eq:dot_prod}:
\begin{multline}
    \mathcal{L}_{h} = \frac{1}{M} \sum_{i=1}^{M} softplus(    \left< \mathcal{H}\left([x^i_k, x^i_{N- k}]\right), h \right>  \\
\times   \left<\mathcal{H}\left([x^i_{N - k}, x^i_k]\right), h\right>),
\end{multline}
where $softplus(t) = \log(1 + e^x)$ is used to give less penalty for correct prediction (when the forward and backward sequences are on different sides of $h$).

In this work, we use a standard residual network \cite{he2016deep} with a fully connected layer so that the output is a vector of length $n$, where $n$ is a hyperparameter of the network. We sample a random hyperplane $h$ in this $n$-dimensional space and fix it during training.

\subsection{Addressing order-ambiguity with HyperCUT}
The function $\mathcal{H}$ can be combined with other losses and used as a regularization to solve the order ambiguity issue as shown in \cref{fig:overall_architecture}b. Specifically, we force the vector corresponding to the output of the deblurring network to lie on only one side of the hyperplane. This can be done by adding to the training loss the following HyperCUT regularization: 
\begin{align}
    \mathcal{R}_{hyp}(f) = \frac{1}{M}\sum_{i=1}^M \sum_{k = 0}^{\left \lfloor{\sfrac{N}{2}}\right \rfloor }\left<\mathcal{H} \left( f_{k}(y^i), f_{N - k}(y^i) \right), h\right>
\end{align}
where $\mathcal{H}$ is the pretrained network described in \cref{sec:hypercut} and it is frozen during the training of deblurring networks. This regularization enforces all synthesized pairs to stay on the ``negative side'' of the hyper-plane $h$. It can be combined with other losses. The final loss for model training will be:
\begin{align}
    \mathcal{L}(f) = \frac{1}{M}\sum_{i=1}^{M}\mathcal{L}_{D}(f(y^i)) + \alpha\mathcal{R}_{hyp}(f),
\end{align}
where $\mathcal{L}_{D}$ can be any $blur2vid$ loss, such as regular $L_2$ loss \cite{zhong2022animation} or order-invariant loss \cite{jin2018learning}, and $\alpha$ is the weight of the HyperCUT regularization. This loss is differential w.r.t. $f$ and can be optimized using any  gradient-based optimizer.

\begin{figure}[t]
    \centering
    \includegraphics[scale=0.205]{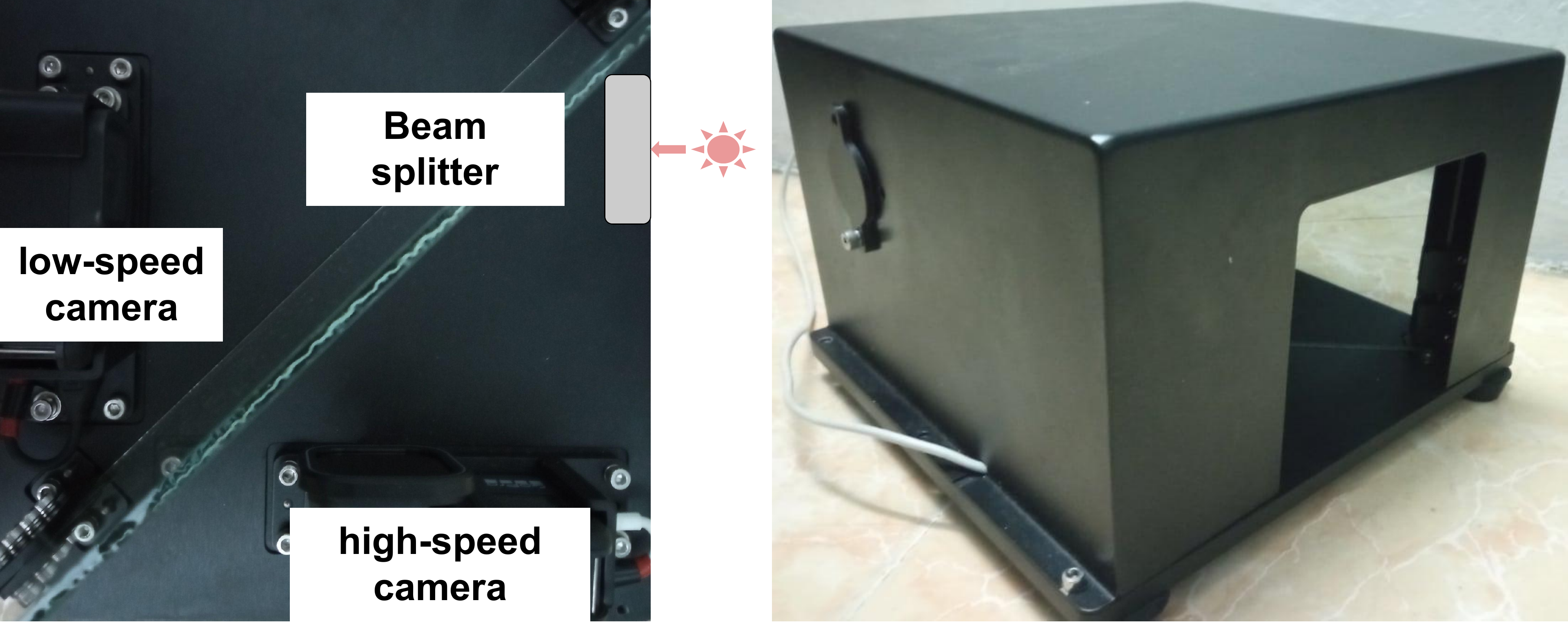}
    \vskip -0.03in
    \caption{Beam splitter camera system\label{ fig:cam_sys}: interior (top-view) and exterior (side-view).}
\end{figure}

\begin{figure}[t]
    \centering
    \includegraphics[scale=0.21]{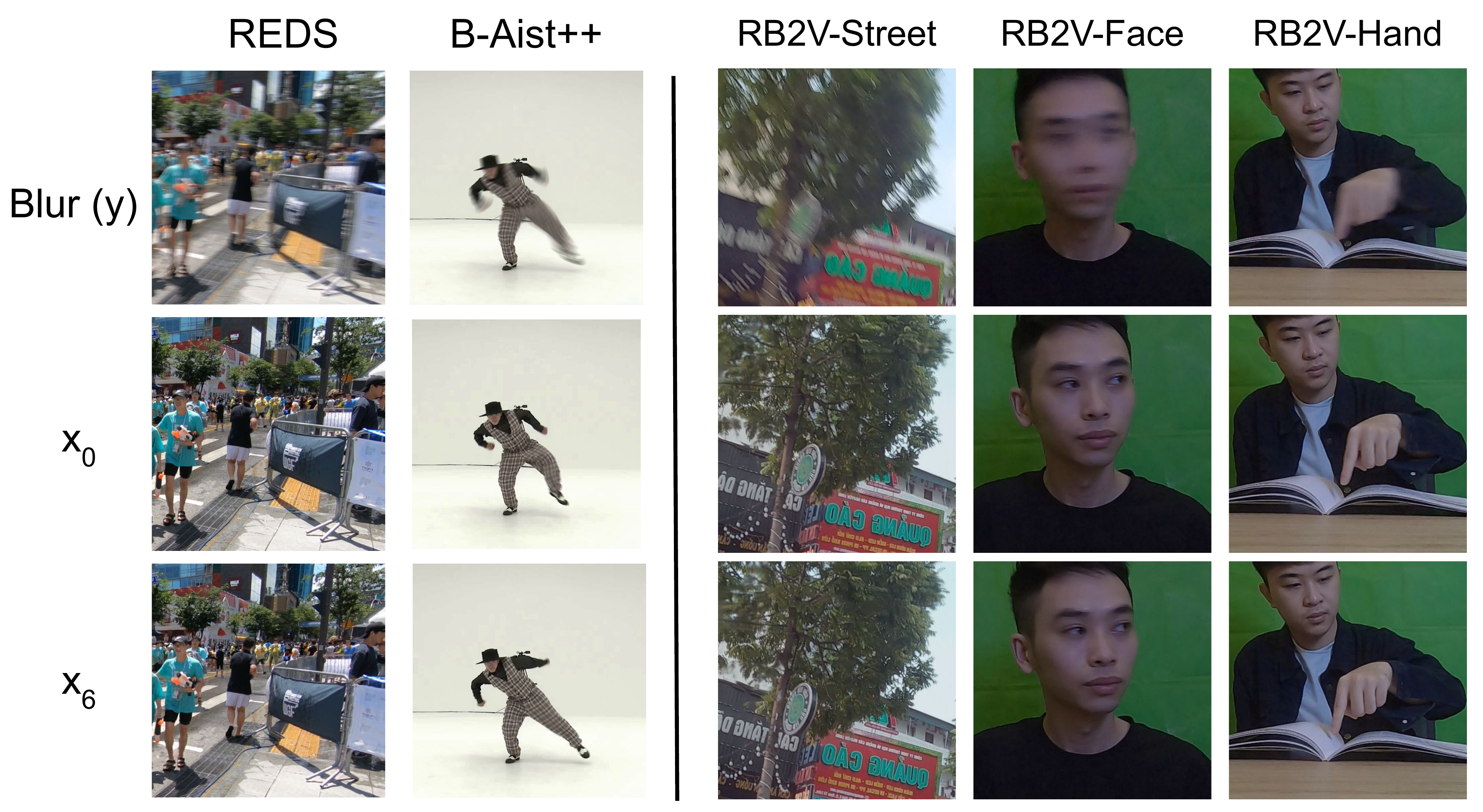}
    \vspace{-2mm}
    
    \caption{Comparing \textit{$blur2vid$} datasets. Sample images from the proposed RB2V dataset are on the right, which are real images as opposed to synthetic ones of existing datasets on the left. \label{fig:data_samples}}
    \vspace{-2mm}
\end{figure}

\section{Real $blur2vid$ (RB2V) Dataset}

Due to the difficulty of collecting paired blurry and sharp video sequences, previous deblurring works utilized synthetic datasets \cite{nah2019ntire,tao2018scale}. Although these datasets are formed by mimicking the camera process (\Eref{blur_formula}), there is a significant gap between synthetic and real blur \cite{tran2021explore}. Recently, beam splitter deblurring datasets \cite{rim2020real,zhong2020efficient} have been proposed and brought remarkable advances in image deblurring. However, such kinds of datasets are not available for the \textit{$blur2vid$} task. This poses the need for the collection of a new dataset for this task. This section describes our data collection procedure.

\subsection{Data collection}

Following recent works~\cite{rim2020real,zhong2021towards,zhong2020efficient}, we built a beam-splitter camera system for collecting real data. This system had two GoPro Hero-8 cameras and a beam splitter as shown in \cref{ fig:cam_sys}. One camera captured videos at 25fps, and the second camera captured at 100fps. 

We used this beam splitter system to collect three categories of data: street, hand, and face, which will be referred to as RB2V-Street, RB2V-Hand and RB2V-Face respectively. For the RB2V-Street dataset, we captured various scenes with different and non-uniform moving objects, creating a diverse dataset with many blur types and intensities. As for RB2V-Face and RB2V-Hand, we captured data for 25 objects in an laboratory environment with a green screen, which was then be replaced with a random static background. For each face or hand object, we captured it with different distances and movements. We divided the datasets into disjoint training and testing sets. \cref{data_statistics} reports the statistics of this dataset, and \cref{fig:data_samples} provides some data samples in the last three columns.

\setlength{\tabcolsep}{15pt}
\begin{table}[t]
    \centering
    \begin{tabular}{lrr}
        \toprule
         \multirow{2}{*}{Data subset}& \multicolumn{2}{c}{\#data samples} \\
         \cmidrule(lr){2-3} \cmidrule(lr){2-3}
          & Train & Test\\
         \midrule
         RB2V-Street & 9000 & 2053\\ 
         RB2V-Face   & 8000 & 2157\\
         RB2V-Hand   & 12000 & 4722\\
         \bottomrule
    \end{tabular}
    \vspace{-2mm}
    \caption{Statistics of our in-the-wild \textit{$blur2vid$} dataset.
    \vspace{-2mm} \label{data_statistics}}
\end{table}
\setlength{\tabcolsep}{4pt}

\subsection{Data processing}

\myheading{Spatial alignment.} Although we tried our best to position the cameras to capture exactly the same scene, there might still be some misalignment between the captured images. To correct for the misalignment, we calibrated the cameras and performed homography mapping. 

\myheading{Temporal upsampling}. The low-speed camera was four times slower than the high-speed one, so each blurry frame corresponded to four sharp ones. Since the previous works typically used seven, we temporally upsampled the frame sequence captured by the 100fps camera by a factor of two. After this interpolation step, each blurry frame corresponded to seven sharp frames, including four original frames and three interpolated ones. We used \cite{ding2021cdfi} as the interpolation module.

\myheading{Color correction and temporal alignment.} Let $C_{x, y}(z)$ denote the color correction algorithm that applies the correction matrix calculated from a reference pair $\{x, y\}$ to an image $z$. Details of this algorithm are given in the supplementary materials. Also denote $y_i^{fake}$ as the synthetic blurry frame generated from the consecutive sharp frames $\{x[i], x[i + 1], ..., x[i + 6]\}$ by temporally upsampling this set to a higher frame rate as in \cite{nah2019ntire} and average all of them. We interpolated two extra frames in between for each consecutive frames, so the number of frames in the upsampled sequence was 19; this helped the synthetic blurry image be more realistic. To find the sharp sequence that corresponded to $y$, denoted as $\mathcal{X} = {x_0, x_1, ..., x_6}$, we needed to find a color correction map $C^{*}$ and a position $p$ so that
\begin{align}
    \mathcal{X} = \{C^{*}(x[p]), C^{*}(x[p+1]), ..., C^{*}(x[p+6])\}
\end{align}
To find $C^{*}$ and $p$, we found the seven consecutive sharp frames such that the ``fake'' blurry image generated by them after color correction was the closest to the real blurry one. If the camera response function $g$ was linear, we have:
\begin{align}
    y &\approx \frac{\sum_{i=0}^{N}C^{*}(x[i])}{N+1}
    = C^{*}  \left(\frac{\sum_{i=0}^{N}(x[i])}{N+1}\right) = C^{*}\left(y_i^{fake}\right). \nonumber 
\end{align}
From the above equation, if we apply $C^{*}$ to $y_i^{fake}$, $y_i^{fake}$ will become $y$. This observation suggests that $C^{*}$ can be approximated by $C_{y_i^{fake}, y}$.

In summary, the position $p$ was found by optimizing:
\begin{align}
    p = \argmin_{i} PSNR\left(C_{y_i^{fake}, y}\left(y_i^{fake}\right), y\right).
\end{align}
The sharp-image sequence $\mathcal{X}$ was taken as the set $\{C_{y_p^{fake}, y}(\{x[p]), C_{y_p^{fake}, y}(\{x[p+1]), ..., C_{y_p^{fake}, y}(\{x[p+6])\}$.
More details are given in the supplementary materials.

\begin{table*}[ht]
    \centering
    \setlength{\tabcolsep}{8pt}
    \setlength{\arrayrulewidth}{0.75pt}
    \begin{tabular}{l|c|cccccccc}
        \toprule
         Model & & ${1^{st}}$ & $2^{nd}$ & ${3^{rd}}$ & $4^{th}$ & ${5^{th}}$ & $6^{th}$ & ${7^{th}}$\\
         
         \hline
        \cite{jin2018learning} & \multirow{4}{*}{\rotatebox[origin=c]{90}{\small{REDS}}} & 20.65 & 22.63 & 24.20 &  23.50 & 24.20 & 22.63 & 20.65 \\
        \cite{jin2018learning} + HyperCUT & & \textbf{22.87} & \textbf{24.88} & \textbf{26.29} & \textbf{25.10}  & \textbf{26.29} & \textbf{24.88} & \textbf{22.86}\\
        \cmidrule{1-1}\cmidrule{3-9}
        \cite{purohit2019bringing} &  & 22.78 & 24.47 & 26.14 & \textbf{31.50} &  26.12 & 24.49 & 22.83\\
        
        \cite{purohit2019bringing} + HyperCUT & & \textbf{26.75} & \textbf{28.30} & \textbf{29.42} & 29.97 & \textbf{29.41} & \textbf{28.30} & \textbf{26.76}\\
        \midrule
         
        \cite{purohit2019bringing} & \multirow{2}{*}{\rotatebox[origin=c]{90}{\small{RB2V}}} & 26.99 & 27.99 & 29.45 &	\textbf{32.08}	&29.55	& 28.06	& 27.04 \\
        \cite{purohit2019bringing} + HyperCUT && \textbf{28.29}	& \textbf{29.20} & \textbf{30.43} & \textbf{32.08} &	\textbf{30.53} & \textbf{29.22} & \textbf{28.25}\\

        \bottomrule
    \end{tabular}
    \vspace{-2mm}
    \caption{pPSNR scores (dB) between predicted frames and the ground-truth ones on the synthetic \textit{$blur2vid$} REDS dataset and our proposed real  $blur2vid$ dataset (we get the average result in all categories including hand, face and street).\label{quan_table}}
    \vspace{-2mm}
\end{table*}

\begin{figure*}[t]
    \centering
    \includegraphics[width=.9\textwidth]{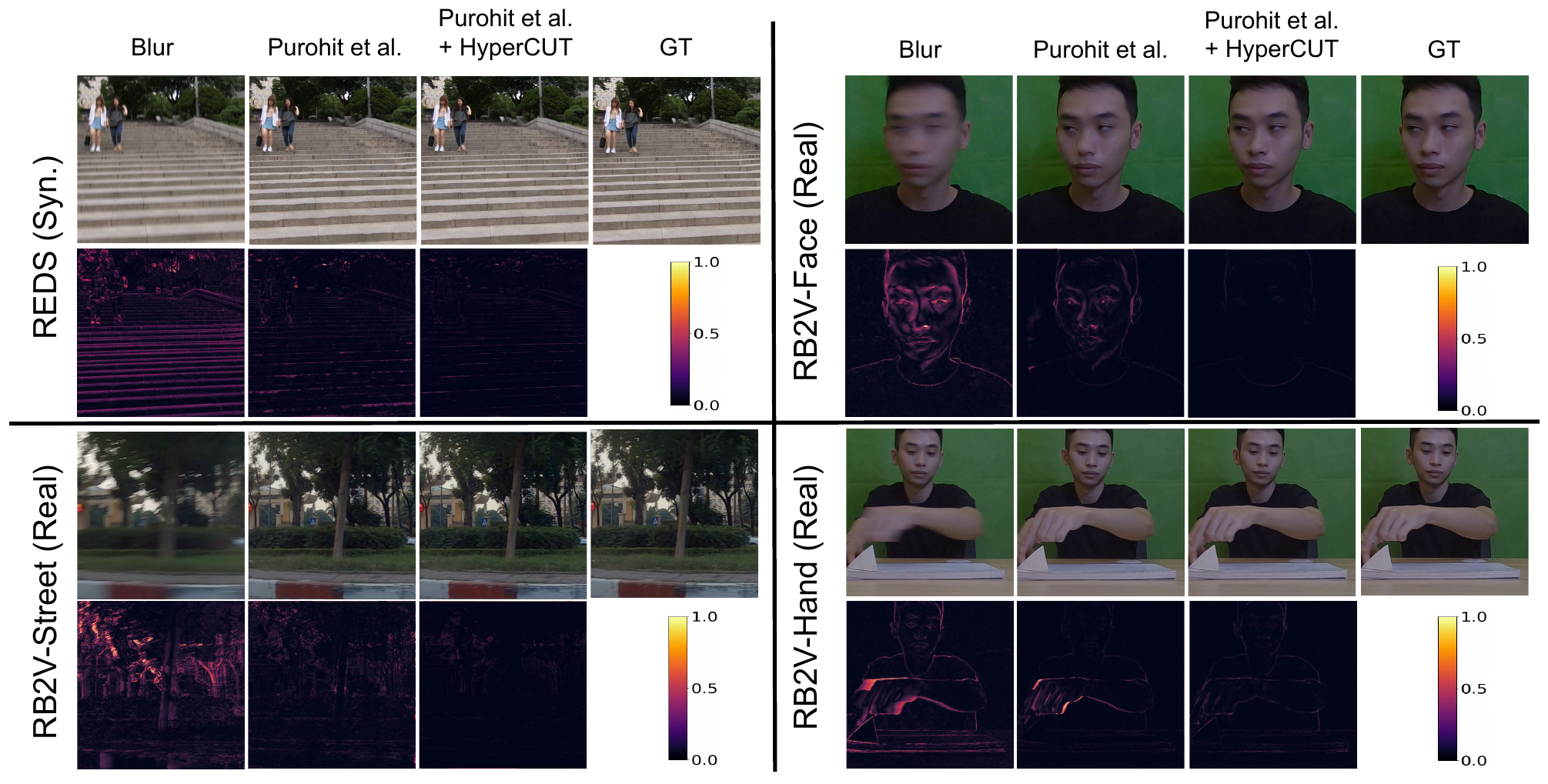}
    \vspace{-2mm}
    \caption{Qualitative results for the first frame prediction on different $blur2vid$ datasets: REDS, RB2V-Street, RB2V-Face, and RB2V-Hand. From left to right on each dataset: blurry input, the result of \cite{purohit2019bringing}, our prediction result applying HyperCUT, and the ground truth. Each result includes a color image and an error heatmap. \label{fig:qual}}
\vspace{-4mm}
\end{figure*}

\section{Experiments}

We compare the proposed ordering scheme applying to the publicly available \textit{$blur2vid$} model proposed by \cite{jin2018learning,purohit2019bringing} on both the existing and the proposed RB2V datasets. In addition, to study the contribution of our proposed mapping, we examine HyperCUT on \cite{zhong2022animation} with the same settings on the B-Aist$++$ dataset \cite{zhang2020mediapipe, li2021ai}. 

\subsection{Dataset preparation}
\minisection {Synthetic datasets.} We used the 120fps set of the REDS dataset \cite{nah2019ntire} to synthesize the training and the testing set (the first row of \cref{fig:data_samples}). Specifically, for every four consecutive frames in the set, we interpolated one intermediate frame between each consecutive pair using CDFI model \cite{teed2020raft}, form a sequence of seven frames, and generated the corresponding blurry image. This allowed us to compare our method with the model proposed by \cite{jin2018learning},  which fixed the number of frames per sequence to seven. 
In addition, we synthesized another testing set using Vimeo90K \cite{xue2019video}. Compared to REDS, this dataset had many more scenes but provided only three frames per data point. Hence, it was suitable for the ablation studies, which required only ground truth on two border frames, but unsuitable for other evaluations. From the three original frames, we interpolated two frames in the middle of each consecutive pair, forming a 7-frame sequence and generating the corresponding blurry image by the same procedure. As for the B-Aist$++$ dataset, we used the augmentation and setting proposed in the original paper that croped the main character using a given bounding box to compare the results from our method and their model. 

\minisection {RB2V dataset.}  We also evaluated the models on our proposed real \textit{$blur2vid$} dataset RB2V on all the three categories. For each category, we re-trained our model and the baselines and tested on the testing set of the same category.

\subsection{Implementation details}
All the models used in the experiments were trained using the Adam optimizer \cite{kingma2014adam}. Training our model took roughly one day for 100 epochs on a single Nvidia A100 GPU. For fair comparison, we re-trained the baseline model on both synthetic and our real datasets.

\subsection{Order Accuracy of HyperCUT}
We trained our HyperCUT model in both synthetic and real datasets to regularize the corresponding $blur2video$ task. In all experiments, we used output vectors of length $n = 128$. For evaluation, we proposed new metrics that overcame the limitations of existing ones in analyzing the effectiveness of our scheme:

$\bullet$ $\mathbold{hit}$: is the ratio of frame pairs $(x_{k}, x_{N-k})$ that satisfy:
\vspace{-4mm}
\begin{align*}
    \left< \mathcal{H}\left([x_k, x_{N - k}]\right), h\right>  \left<\mathcal{H}\left([x_{N - k}, x_k]\right), h\right> < 0
\end{align*}

$\bullet$ $\mathbold{con}$: measures the \textbf{con}sistency of frame pairs in each sequence in the HyperCUT space. It computes the ratio that the pairs ($x_1$, $x_7$), ($x_2$, $x_6$), and ($x_3$, $x_5$) are in the same side of the hyperplane $h$.

\begin{figure}[!htb]
\centering
\begin{subfigure}{0.23\textwidth}
    \centering
    \includegraphics[width=\linewidth]{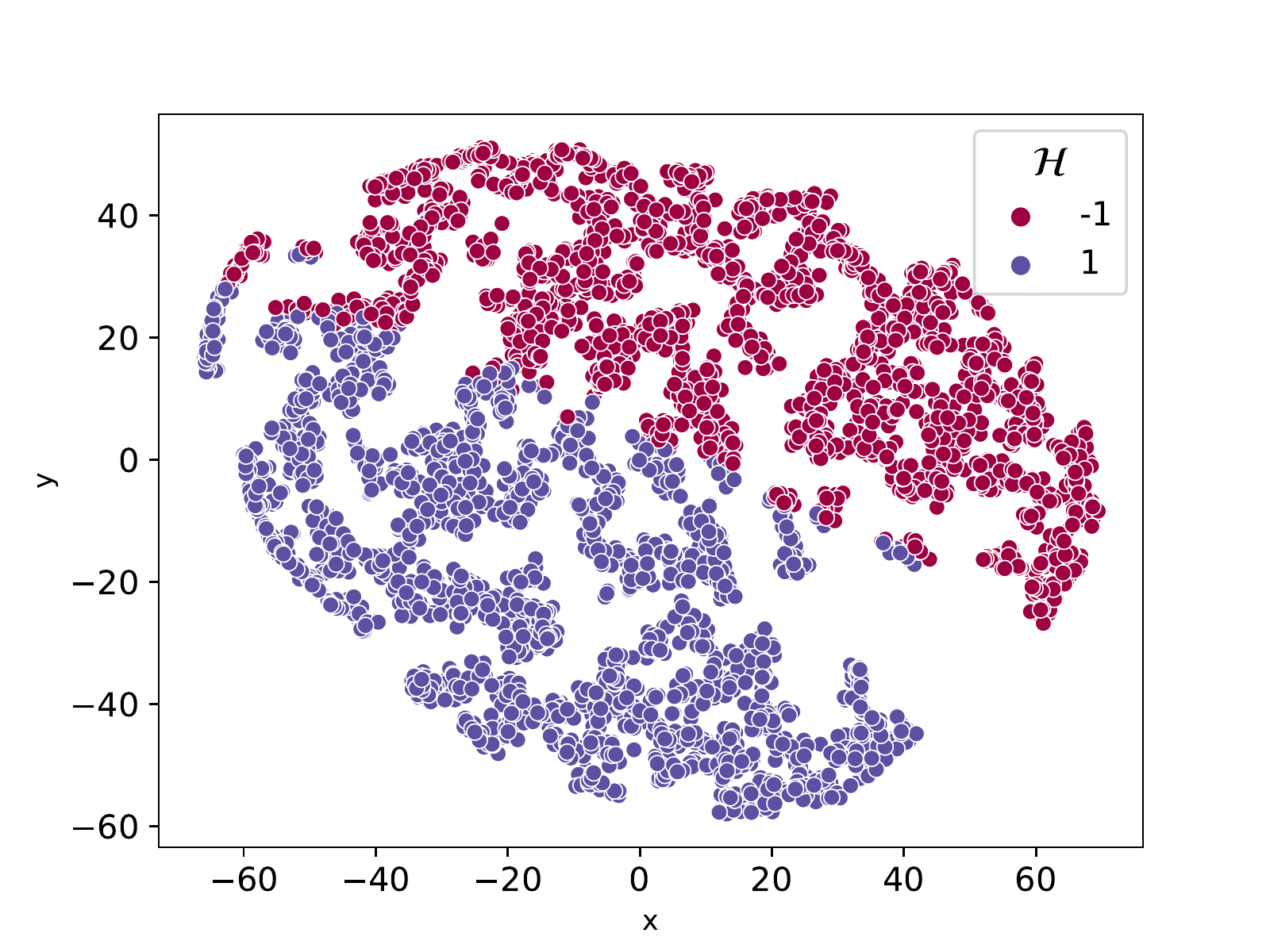}
    \caption{}
    \label{fig:tsne_street}
\end{subfigure}%
\begin{subfigure}{0.23\textwidth}
    \centering
    \includegraphics[width=\linewidth]{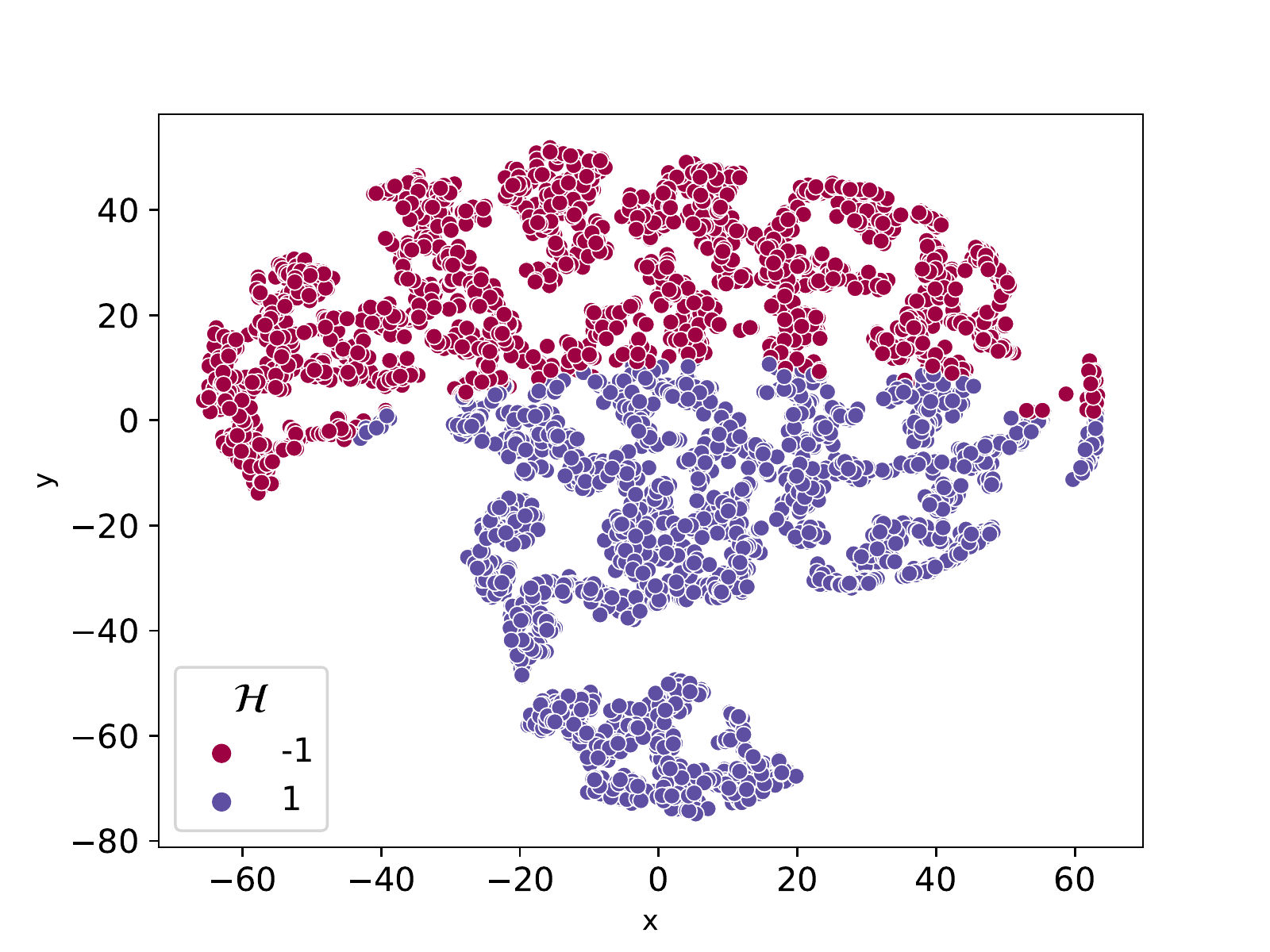}
    \caption{}
    \label{fig:tsne_reds}
\end{subfigure}
\vskip -0.05in
    \vspace{-2mm}
\caption{The t-SNE visualization of HyperCUT ordering mapping on (a) RV2B-Street (Real) and (b) REDS (Synthetic) datasets. We use $-1$ and $1$ to represent each side of hyperplane. } 
\vspace{-4mm}
\label{fig:tsne}
\end{figure}

As can be seen in \cref{tab:hypercut}, our proposed self-supervised model can extract the ordering information effectively in all mentioned datasets. The trained models achieve almost perfect scores in all metrics. We use t-SNE to visualize representation vectors in \cref{fig:tsne}. It can be observed that the mapped vectors are perfectly split into two clusters in both REDS and RV2B-Street datasets.

\begin{table}[ht]
    \centering
\setlength{\tabcolsep}{10pt}
    \begin{tabular}{lccc}
    \hline
    Dataset   & $hit$ & $con$@$2$ & $con$@$3$ \\
    \hline 
    REDS & 95.7 & 96.5 & 94.4 \\
    B-Aist$++$ & 97.5 & 95.6 & 91.2 \\
    \hline
    RB2V-Face & 94.4 & 96.7 & 92.1 \\
    RB2V-Hand & 98.6 & 97.0 & 96.7 \\
    RB2V-Street & 98.7 & 98.3 & 96.8 \\
    \hline
    \end{tabular}
    \vspace{-2mm}
    \caption{The experiments of HyperCUT on five datasets with $hit$ (\%) and $con$ (\%) rates. We denote $con$@$X$ as the consistency rate of HyperCUT ordering for two frame pairs $(x_i, x_{N-i}), (x_j, x_{N-j})$ (when X = 2) and for 3 frame pairs $(x_i, x_{N-i}), (x_j, x_{N-j}), (x_k, x_{N-k})$ (when X = 3) in the same side of the hyperplane.}
\vspace{-4mm}
    \label{tab:hypercut}
\end{table}
\subsection{HyperCUT Regularization}
\vspace{-2mm}
\minisection{Synthetic datasets.} We studied the HyperCUT regularization with the methods proposed by \cite{jin2018learning,purohit2019bringing} and \cite{zhong2022animation} on the REDS and B-Aist$++$ datasets, using a default weight $\alpha = 0.2$.

\begin{table*}[htb]
\centering
\begin{tabular}{cccc}
\toprule
Method & \cite{zhong2022animation} (from paper) &  \cite{zhong2022animation} (reproduced)  & \cite{zhong2022animation} + HyperCUT \\
\midrule
 $\mathcal{P}_1$ & \textcolor{RedViolet}{$19.97$} /  \textcolor{teal}{$0.860$} /  \textcolor{Blue}{$0.089$} & \textcolor{RedViolet}{$20.58$} / \textcolor{teal}{$0.890$} / \textcolor{Blue}{$0.068$} & \textcolor{RedViolet}{$22.16$} /  \textcolor{teal}{$0.901$} / \textcolor{Blue}{$0.102$} \\

$\mathcal{P}_3$ & \textcolor{RedViolet}{$22.44$} /  \textcolor{teal}{$0.890$} /  \textcolor{Blue}{$0.068$} & \textcolor{RedViolet}{$21.21$} / \textcolor{teal}{$0.899$} / \textcolor{Blue}{$0.063$} & \textcolor{RedViolet}{$23.31$} /  \textcolor{teal}{$0.915$} / \textcolor{Blue}{$0.062$} \\

$\mathcal{P}_5$ & \textcolor{RedViolet}{$23.49$} /  \textcolor{teal}{$0.911$} /  \textcolor{Blue}{$0.060$} & \textcolor{RedViolet}{$22.48$} / \textcolor{teal}{$0.903$} / \textcolor{Blue}{$0.061$} & \textcolor{RedViolet}{$23.81$} /  \textcolor{teal}{$0.920$} / \textcolor{Blue}{$0.060$} \\

\bottomrule
\end{tabular}

\vspace{-2mm}
\caption{\textbf{Quantitative evaluation of the blurry image decomposition}. \textcolor{RedViolet}{$p\overline{\text{PSNR}}$ $\uparrow$} , \textcolor{teal}{$p\overline{\text{SSIM}}$ $\uparrow$}, and \textcolor{Blue}{$p\overline{\text{LPIPS}}$ $\downarrow$} are used as evaluation metrics. For Zhihang et al. \cite{zhong2022animation}, we predict multiple motion guidance from the guidance predictor network. $\mathcal{P}_{\#}$ denotes we evaluate $\#$ number of plausible decomposition results for each input, and choose the best. The results of Zhihang et al. \cite{zhong2022animation} with the HyperCUT regularization represent the  best performance calculated using either the forward or reverse outputs, following the original paper.}
\label{tab:animation}
\vspace{-2mm}
\end{table*}

With the REDS dataset, we re-trained the models proposed in \cite{jin2018learning} and \cite{purohit2019bringing}  with their original loss functions and with our proposed HyperCUT add-on. For evaluation, since $\mathcal{L}_{OI}$ accepts any frame ordering, we define a paired-based PSNR, denoted as pPSNR, that computes the maximum average of PSNR scores between the regressed and ground-truth symmetric frame pair in forward and backward order. Specifically, given the output $(x'_0, x'_1, ..., x'_N)$ and the ground-truth $(x_0, x_1, ..., x_N)$, pPSNR can be computed as:

\vspace{-5mm}
\begin{multline}
pPSNR_{k}(x, x') = max(PSNR(x'_k, x_k),\\
PSNR(x'_k, x_{N - k}))
\label{psnr}
\end{multline}

Quantitative results are given in \cref{quan_table}, where we use pPSNR scores to measure the performance of the models. Our HyperCUT-based models provide stable performance on all frames and consistently outperform the compared models on all six border frames with 1-4 point gaps in pPSNR scores. Since the backbone used in \cite{jin2018learning} is weak and outdated, from now on, we will focus on the models with the more recent and stronger backbone of \cite{purohit2019bringing}. We notice that the performance gap caused by HyperCUT regularization increases when moving to the boundary frames $x_0$ and $x_7$. The compared model, however, performs better at the center frame with an exceptionally high {pPSNR}. This model performs poorly on border frames, meaning that its loss concentrates on improving the quality of the middle frame. Our model, on the other hand, has a balance in improving all frames together. While our pPSNR score on the middle frame is not as high, we can easily improve it by deploying an extra, normal image deblurring network. The border frames, on the other hand, can only be learned effectively with our proposed HyperCUT regularization. An example is shown in the top left of \cref{fig:qual}. The result produced by our model is sharper and closer to the ground truth.

In addition, with the benchmark proposed by Zhong \etal \cite{zhong2022animation} on the B-Aist$++$ dataset, we re-train the model with the same setting as the original model for a fair comparison, evaluating by average pPSNR, pSSIM, and pLPIPS\cite{zhang2018unreasonable} metrics, in which pSSIM and pLPIPS are defined similar to pPSNR. We denote these metrics as $p\overline{\text{PSNR}}$, $p\overline{\text{SSIM}}$, and $p\overline{\text{LPIPS}}$, respectively. As can be seen in \cref{tab:animation}, the result with HyperCUT regularization dominates the one reported from the paper as well as the reproduced version. The score difference between the two versions of the original method also reveals the instability of the motion guidance module.

\minisection{RB2V dataset.}
We also ran evaluation on our proposed RB2V dataset. We trained the models using the training set of each category and then test on the corresponding test set. The quantitative and qualitative results are given in \cref{quan_table} and the bottom left of \cref{fig:qual}, respectively. Again, our model shows better accuracy and reconstruction quality than~\cite{purohit2019bringing}.

\minisection{Applications of $blur2vid$ for faces and hands.}
We tested the models on domain-specific datasets and measured the ability of recovering face and hand trajectories from a single blurry image. 
Given a blurry face image, we first run a $blur2vid$ model to obtain a sequence of sharp images, each of which would  subsequently be fed into a facial landmark detection algorithm to detect 68 facial landmarks. To measure the quality of a recovered face trajectory, we calculated the Mean Squared Error (MSE) between the 68 facial landmarks detected on the recovered sharp image and the 68 facial landmarks detected on the ground truth sharp image. Similarly for hands, we detected the tip of the index finger in each recovered sharp image using the hand detection algorithm \cite{zhang2020mediapipe}, and calculated its distance to the index finger detected on the ground truth sharp image. 

Quantitative results of face and hand trajectory recovery are given in \cref{quan_trajec}. Compared to the baseline, the proposed model with HyperCUT regularization  was more accurate, with reasonably small error for practical applications.

\begin{table}[t]
    \centering
    \begin{tabular}{lrr}
        \toprule
         Deblur method & Face & Hand\\
        \midrule
      Purohit \etal    \cite{purohit2019bringing} & 5.75 & 11.67\\
        Purohit \etal    \cite{purohit2019bringing} + HyperCUT  & \textbf{4.87} & \textbf{9.2}\\
         \bottomrule
    \end{tabular}
\vspace{-3mm}
    \caption{Quantitative results for face and hand trajectory recovery from a single blurry image. \label{quan_trajec}}
    \vspace{-4mm}
\end{table}
\vspace{-2mm}

\subsection{Ablation Studies}
\vspace{-2mm}
\myheading{Regularization weight for HyperCUT.} We ablated the weight parameter $\alpha$ to have a deeper understanding of its effect on the final performance. We experimented with different values for $\alpha$ from $0$ to $0.3$ with Purohit \etal \cite{purohit2019bringing} backbone on the RB2V-Street dataset, and computed the mean pPSNR score, denoted as $\overline{\text{pPSNR}}$. The results are reported in \cref{tab:alpha}. As can be seen, when $\alpha$ was increased, the $\overline{\text{pPSNR}}$ score gradually increased and peaked at $\alpha {=}0.2$, confirming the positive contribution of the HyperCUT loss. When $\alpha{>}0.2$, the ordering information started to outweigh the order-invariant loss, decreasing the score. Hence, we selected  $\alpha = 0.2$ as the default setting for other experiments. 
\vspace{-7mm}
\begin{table}[H]
\resizebox{\columnwidth}{!}{
\begin{tabular}{c|cccccc}
$\alpha$ & 0 & 0.1  & 0.15  & 0.2   & 0.25 & 0.3   \\ 
\midrule 
$\overline{\text{pPSNR}}$ (dB) & 25.73 & 25.8 & 26.33 & \textbf{26.95} & 26.9 & 25.75\\
\end{tabular}
}
\vspace{-3mm}
\caption{The $\overline{\text{pPSNR}}$ (dB) of seven generated sharp frames on the RB2V-Street dataset when changing $\alpha$.}
\label{tab:alpha}
\vspace{-3mm}
\end{table}

\vspace{-3mm}
\minisection{Dimension of the hyperplane.} We experimented with different settings for the number of dimensions of the hyperplane, and \cref{tab:hyper} shows the hit and consistency ratios of HyperCUT on the RB2V-Street and RB2V-Hand datasets. As can be seen, with $n \geq 16$, the accuracy of the order assigning is consistent with a small variance. In most of our experiments, we used $n = 128$ due to its best overall performance in terms of hits and cons ratios.

\vspace{-2mm}
\begin{table}[ht]
\centering
\begin{tabular}{|c|ccc|ccc|}
\hline
\multirow{2}{*}{$n$} & \multicolumn{3}{|c|}{RB2V-Street} & \multicolumn{3}{|c|}{RB2V-Hand}\\
\cline{2-7}
 &$hit$ & $con$@$2$ & $con$@$3$ &$hit$ & $con$@$2$ & $con$@$3$ \\ \hline
1     & 95.5 & 97.7  & 94.5  & 95.0 & 96.3  & 92.5\\
16    & 98.4 & 98.2  & 96.1  & 96.8 & 96.6  & 92.8\\
64    & \textbf{99.1} & 98.4  & 96.4  & 97.4 & 96.0  & 93.5\\
128   & 98.7 & \textbf{98.5}  & \textbf{96.8} & \textbf{98.6} & \textbf{97.0}  & \textbf{96.7}\\
256   & 97.5 & 97.7  & 95.2  & 96.9 & 96.8  & 93.3\\ \hline
\end{tabular}
\vspace{-2mm}
\caption{Ablation study for $n$, the dimension of the HyperCUT hyperplane.}
\label{tab:hyper}
\vspace{-3mm}
\end{table}

\vspace{-3mm}
\section{Conclusions}

\vspace{-2mm}
In this paper, we have proposed a method for the $blur2vid$ task, effectively addressing the order-ambiguity issue with an innovative regularization called HyperCUT. The regularization assigns an order label to each potential solution and enforces the $blur2vid$ model to generate only that specific solution, thereby enhancing its performance. The proposed regularization can be implemented with any existing $blur2vid$ model for substantial improvements. Furthermore, we contributed a novel dataset for the development and evaluation of the image-to-video deblurring task. This dataset comprises real images from three distinct domains, namely street, face, and hand.

In this work, we focus on standard motion blur in normal capturing conditions with short exposure time, resulting in simple and consistent direction and velocity. Future research on adapting HyperCUT for handling complex movements and long exposure blur would be an interesting avenue for exploration.

{\small
\bibliographystyle{ieee_fullname}
\bibliography{longstrings,egbib}
}
\end{document}


\input{definitions}

\title{HyperCUT: Video Sequence from a Single Blurry Image \\ using Unsupervised Ordering 
      \\---~Supplementary Material~---}

\author{
Bang-Dang Pham$^{1,*}$ \quad Phong Tran$^{1,2,*}$ \quad Anh Tran$^{1}$ \quad Cuong Pham$^{1,3}$ \quad Rang Nguyen$^{1}$\quad Minh Hoai$^{1,4}$\\ 
\small{\textsuperscript{1}VinAI Research, Vietnam \quad \textsuperscript{2}MBZUAI, UAE \quad \textsuperscript{3}Posts \& Telecommunications Inst. of Tech., Vietnam
\quad \textsuperscript{4}Stony Brook University, USA}\\
\texttt{\scriptsize \{v.dangpb1, v.anhtt152,  v.hoainm\}@vinai.io} \quad 
\texttt{\scriptsize cuongpv@ptit.edu.vn   }
\texttt{\scriptsize  the.tran@mbzuai.ac.ae} \\
\small{\textsuperscript{*}Equal contribution}
}
\maketitle

\begin{abstract}
   In this supplementary PDF, we first specify some techniques applied to our data post-processing. Then, we provide the details of our proposed HyperCUT architecture. Finally, we illustrate and analyze the qualitative results compared with the SOTA model. 
\end{abstract}

\section{Data post-processing }
\begin{figure*}
    \centering
    \includegraphics[scale=0.25]{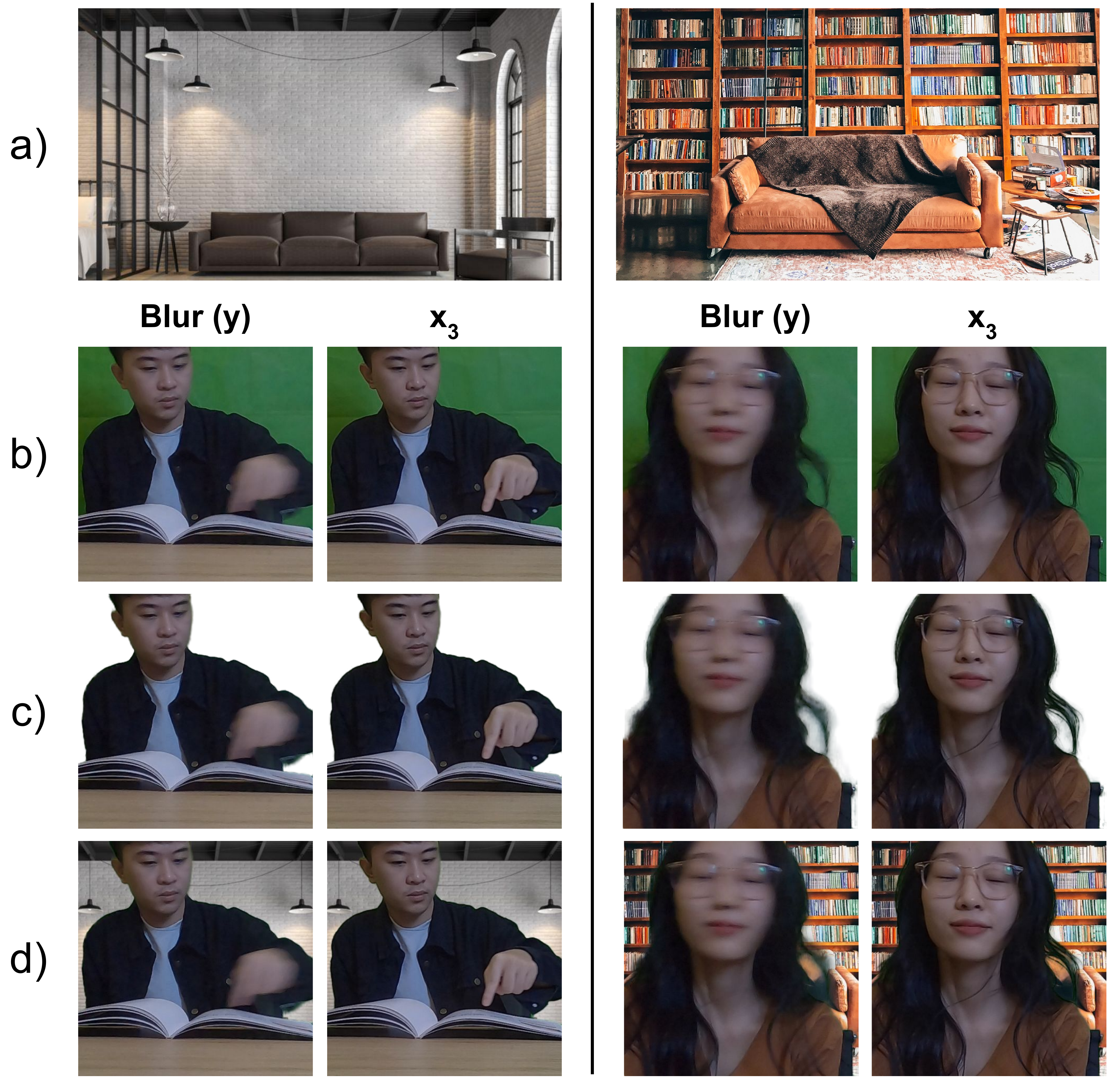}
    \caption{\textbf{Matting examples}. Row (a) are random background images collected on the Internet. Row(b) are images captured using green screen. Row (c) are extracted foreground using \cite{xue2019video2}. Row (d) are the final images used to train the models.}
    \label{fig:matting}
\end{figure*}

\minisection{Random background for Face/Hand subsets.} For the datasets collected using green screen, we use \cite{lin2021real2} to extract the foreground objects and then blend them into random backgrounds collected on the internet. Some examples are shown in \Fref{fig:matting}.

\minisection{Color correction algorithm.} Given two reference images ($x, y$), we calculate the color correction matrix of the pair (The map $C_{x, y}(\cdot)$ in the paper) by minimizing:
\begin{align}
    M^* = \argmin_{M \in R^{3x4}} \|Mx - y\|
\end{align}
$M^*$ can be easily calculated by using linear regression. 

Equation 11 in the paper explains how we can find reference images to calibrate color between two cameras. To find the frame index $p$ in that equation, we simply iterate all possible position, and then choose the one $i$ with largest $PSNR(C_{y_i^{fake}}, y)$. Details are given in \Aref{algo:calibration}.

\begin{algorithm}[ht]
    \caption{Color correction}
    \label {algo:calibration}
    
    \textbf{Input:} $y$, $x[0], x[1], ..., x[h]$\\
    \textbf{Output:} 7 calibrated sharp frames $x_0, x_1, ..., x_6$
    
    \begin{algorithmic}[1]
        \State $i \leftarrow 0$
        \State $p \leftarrow -1$
        \State $best \leftarrow -\infty$
        \While {$i + 6 < h$}
        
            \State $y_{fake} \leftarrow synBlur(x[i], x[i + 1], ..., x[i + 6])$
            
            \State $C \leftarrow ColorCorrectionMap(y_{fake}, y)$
            
            \If {$PSNR(C(y_{fake}), y) > best$}
                \State $p \leftarrow i$
            \EndIf
            
            \State $i \leftarrow i + 1$
        \EndWhile
    
        \State $x_0, x_1, ..., x_6 = x[p], x[p + 1], ..., x[p + 6]$
    \end{algorithmic}
\end{algorithm}

In addition, for the datasets collected in the laboratory (face and hand categories of our proposed dataset), before applying the color correction algorithm mentioned in the paper, we first calibrate the colors of the two cameras using a color checker.

\minisection{Others.} Due to the design of the camera system, after capturing images, we crop the border of the images to remove the black region of the cameras, letting the size of each image be $448 \times 448$.

\subsection{Additional dataset statistics}
In the paper, we use a synthetic blur2vid dataset generated from REDS \cite{nah2019ntire2} for training and testing. Here we provide the numbers of training and testing sequences of the set in \Tref{syn_data_statistics}.

\setlength{\tabcolsep}{10pt}
\begin{table}[ht]
    \centering
    \caption{Statistics of the synthetic datasets used in the paper \label{syn_data_statistics}}
    \vskip 0.1in
    \begin{tabular}{lcc}
        \toprule
         \multirow{2}{*}{Dataset}& \multicolumn{2}{c}{\#data samples} \\
         \cmidrule(lr){2-3} \cmidrule(lr){2-3}
          & Train & Test\\
         \midrule
         REDS & 58876 & 1330 \\ 
         \bottomrule
    \end{tabular}

\end{table}
\setlength{\tabcolsep}{4pt}

\section{Details of the network architecture}
\begin{table}[ht]
    \centering
    \vspace{0.1in}
    \begin{tabular}{|c|c|}
        \hline
        Layer & Output shape \\
        \hline
                             & $H    \times W    \times 6$\\
                             
        \hline
        ReflectionPad2d(3)   & \multirow{2}{*}{$H    \times W    \times 64$}\\
        Conv2d(6, 32, 7, 1, 0)     & \\
        
        \hline
        Conv2d(32, 32, 3, 2, 1)    & \multirow{2}{*}{$\sfrac{H}{2}    \times \sfrac{W}{2}    \times 32$}\\
        LeakyReLU()                 & \\

        \hline
        Conv2d(32, 64, 3, 2, 1)    & \multirow{2}{*}{$\sfrac{H}{4}    \times \sfrac{W}{4}    \times 64$}\\
        LeakyReLU()                 & \\

        \hline
        Conv2d(64, 128, 3, 2, 1)    & \multirow{2}{*}{$\sfrac{H}{8}    \times \sfrac{W}{8}    \times 128$}\\
        LeakyReLU()                 & \\

        \hline
        Conv2d(128, 128, 3, 2, 1)    & \multirow{2}{*}{$\sfrac{H}{16}    \times \sfrac{W}{16}    \times 128$}\\
        LeakyReLU()                 & \\

        \hline
        Conv2d(128, 128, 3, 2, 1)    & \multirow{2}{*}{$\sfrac{H}{32}    \times \sfrac{W}{32}    \times 128$}\\
        LeakyReLU()                 & \\

        \hline
        ResBlock(128, 128) $\times 6$  & $\sfrac{H}{32}    \times \sfrac{W}{32}    \times 128$\\

        \hline
        AdapativeAvgPool2d(1, 1) & $128$\\
        \hline
    \end{tabular}
    \caption{Network architecture}
    \label{tab:f_func}
\end{table}

The detailed architecture of our proposed network for $\mathcal{H}$ is shown in Table \ref{tab:f_func} with $n = 128$. 


\begin{figure*}[ht]
    \centering
    \includegraphics[width=\linewidth]{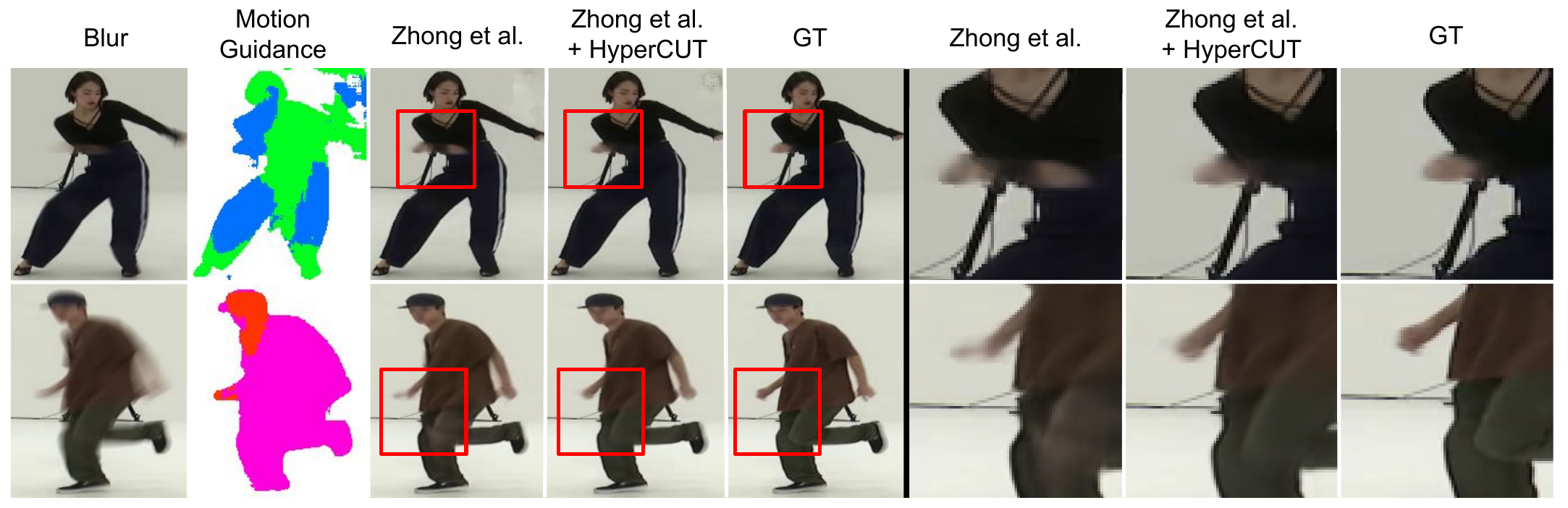}
    \caption{\textbf{Qualitative comparison with \cite{zhong2022animation2} in solving order-ambiguity.} We test Zhong et al.\cite{zhong2022animation2} models with the original setting and after embedding our regularization. The \textcolor{Red}{red} region emphasizes the contribution of our HyperCUT module in overcoming the order-ambiguity issue of the model. For a fair comparison, we use the same motion guidance but predict with two different decomposer modules, one with the original loss\cite{zhong2022animation2} and one with HyperCUT regularization.  }
    \label{fig:red}
\end{figure*}
\section{Additional Qualitative Results}
\vspace{-1mm}
\subsection{Order-Ambiguity Impact} 
We test our embedding module on the Synthetic dataset B-Aist$++$ as mentioned in \cite{zhong2022animation2}. As shown in Fig. \ref{fig:red}, the ordering proposed by our HyperCUT module helps the baseline model overcome the reconstruction issue to some extent when using the same motion guidance. Especially when blur is caused by fast movement, as given in the first row of Fig \ref{fig:red}, the specific direction in the training state through HyperCUT regularization can improve the model stability.

\subsection{Video Result}
We present the video results as \textcolor{red}{result.mp4} in our Github link. We first show an blurry image, then same sample from motion guidance prediction network, the original result of \cite{zhong2022animation2}, result after embedding our HyperCUT module and the corresponding ground truth.


{\small

\bibliographystyle{ieee_fullname}
\bibliography{longstrings,supp}
}

%% file: definitions.tex
\def\mA{\mathcal{A}}
\def\mB{\mathcal{B}}
\def\mC{\mathcal{C}}
\def\mD{\mathcal{D}}
\def\mE{\mathcal{E}}
\def\mF{\mathcal{F}}
\def\mG{\mathcal{G}}
\def\mH{\mathcal{H}}
\def\mI{\mathcal{I}}
\def\mJ{\mathcal{J}}
\def\mK{\mathcal{K}}
\def\mL{\mathcal{L}}
\def\mM{\mathcal{M}}
\def\mN{\mathcal{N}}
\def\mO{\mathcal{O}}
\def\mP{\mathcal{P}}
\def\mQ{\mathcal{Q}}
\def\mR{\mathcal{R}}
\def\mS{\mathcal{S}}
\def\mT{\mathcal{T}}
\def\mU{\mathcal{U}}
\def\mV{\mathcal{V}}
\def\mW{\mathcal{W}}
\def\mX{\mathcal{X}}
\def\mY{\mathcal{Y}}
\def\mZ{\mathcal{Z}} 

\def\bbN{\mathbb{N}} 
\def\bbR{\mathbb{R}} 
\def\bbP{\mathbb{P}} 
\def\bbQ{\mathbb{Q}} 
\def\bbE{\mathbb{E}}

\def\1n{\mathbf{1}_n}
\def\0{\mathbf{0}}
\def\1{\mathbf{1}}

\def\A{{\bf A}}
\def\B{{\bf B}}
\def\C{{\bf C}}
\def\D{{\bf D}}
\def\E{{\bf E}}
\def\F{{\bf F}}
\def\G{{\bf G}}
\def\H{{\bf H}}
\def\I{{\bf I}}
\def\J{{\bf J}}
\def\K{{\bf K}}
\def\L{{\bf L}}
\def\M{{\bf M}}
\def\N{{\bf N}}
\def\O{{\bf O}}
\def\P{{\bf P}}
\def\Q{{\bf Q}}
\def\R{{\bf R}}
\def\S{{\bf S}}
\def\T{{\bf T}}
\def\U{{\bf U}}
\def\V{{\bf V}}
\def\W{{\bf W}}
\def\X{{\bf X}}
\def\Y{{\bf Y}}
\def\Z{{\bf Z}}

\def\a{{\bf a}}
\def\b{{\bf b}}
\def\c{{\bf c}}
\def\d{{\bf d}}
\def\e{{\bf e}}
\def\f{{\bf f}}
\def\g{{\bf g}}
\def\h{{\bf h}}
\def\i{{\bf i}}
\def\j{{\bf j}}
\def\k{{\bf k}}
\def\l{{\bf l}}
\def\m{{\bf m}}
\def\n{{\bf n}}
\def\o{{\bf o}}
\def\p{{\bf p}}
\def\q{{\bf q}}
\def\r{{\bf r}}
\def\s{{\bf s}}
\def\t{{\bf t}}
\def\u{{\bf u}}
\def\v{{\bf v}}
\def\w{{\bf w}}
\def\x{{\bf x}}
\def\y{{\bf y}}
\def\z{{\bf z}}

\def\balpha{\mbox{\boldmath{$\alpha$}}}
\def\bbeta{\mbox{\boldmath{$\beta$}}}
\def\bdelta{\mbox{\boldmath{$\delta$}}}
\def\bgamma{\mbox{\boldmath{$\gamma$}}}
\def\blambda{\mbox{\boldmath{$\lambda$}}}
\def\bsigma{\mbox{\boldmath{$\sigma$}}}
\def\btheta{\mbox{\boldmath{$\theta$}}}
\def\bomega{\mbox{\boldmath{$\omega$}}}
\def\bxi{\mbox{\boldmath{$\xi$}}}
\def\bnu{\mbox{\boldmath{$\nu$}}}                                  
\def\bphi{\mbox{\boldmath{$\phi$}}}
\def\bmu{\mbox{\boldmath{$\mu$}}}

\def\bDelta{\mbox{\boldmath{$\Delta$}}}
\def\bOmega{\mbox{\boldmath{$\Omega$}}}
\def\bPhi{\mbox{\boldmath{$\Phi$}}}
\def\bLambda{\mbox{\boldmath{$\Lambda$}}}
\def\bSigma{\mbox{\boldmath{$\Sigma$}}}
\def\bGamma{\mbox{\boldmath{$\Gamma$}}}
                                  
\newcommand{\myprob}[1]{\mathop{\mathbb{P}}_{#1}}

\newcommand{\myexp}[1]{\mathop{\mathbb{E}}_{#1}}

\newcommand{\mydelta}[1]{1_{#1}}

\newcommand{\myminimum}[1]{\mathop{\textrm{minimum}}_{#1}}
\newcommand{\mymaximum}[1]{\mathop{\textrm{maximum}}_{#1}}    
\newcommand{\mymin}[1]{\mathop{\textrm{minimize}}_{#1}}
\newcommand{\mymax}[1]{\mathop{\textrm{maximize}}_{#1}}
\newcommand{\mymins}[1]{\mathop{\textrm{min.}}_{#1}}
\newcommand{\mymaxs}[1]{\mathop{\textrm{max.}}_{#1}}  
\newcommand{\myargmin}[1]{\mathop{\textrm{argmin}}_{#1}} 
\newcommand{\myargmax}[1]{\mathop{\textrm{argmax}}_{#1}} 
\newcommand{\myst}{\textrm{s.t. }}

\newcommand{\denselist}{\itemsep -1pt}
\newcommand{\sparselist}{\itemsep 1pt}

\definecolor{pink}{rgb}{0.9,0.5,0.5}
\definecolor{purple}{rgb}{0.5, 0.4, 0.8}   
\definecolor{gray}{rgb}{0.3, 0.3, 0.3}
\definecolor{mygreen}{rgb}{0.2, 0.6, 0.2}

\newcommand{\cyan}[1]{\textcolor{cyan}{#1}}
\newcommand{\red}[1]{\textcolor{red}{#1}}  
\newcommand{\blue}[1]{\textcolor{blue}{#1}}
\newcommand{\magenta}[1]{\textcolor{magenta}{#1}}
\newcommand{\pink}[1]{\textcolor{pink}{#1}}
\newcommand{\green}[1]{\textcolor{green}{#1}} 
\newcommand{\gray}[1]{\textcolor{gray}{#1}}    
\newcommand{\mygreen}[1]{\textcolor{mygreen}{#1}}    
\newcommand{\purple}[1]{\textcolor{purple}{#1}}       

\definecolor{greena}{rgb}{0.4, 0.5, 0.1}
\newcommand{\greena}[1]{\textcolor{greena}{#1}}

\definecolor{bluea}{rgb}{0, 0.4, 0.6}
\newcommand{\bluea}[1]{\textcolor{bluea}{#1}}
\definecolor{reda}{rgb}{0.6, 0.2, 0.1}
\newcommand{\reda}[1]{\textcolor{reda}{#1}}

\def\changemargin#1#2{\list{}{\rightmargin#2\leftmargin#1}\item[]}
\let\endchangemargin=\endlist
                                               
\newcommand{\cm}[1]{}

\newcommand{\mhoai}[1]{{\color{purple}\textbf{[MH: #1]}}}

\newcommand{\mtodo}[1]{{\color{red}$\blacksquare$\textbf{[TODO: #1]}}}
\newcommand{\myheading}[1]{\vspace{1ex}\noindent \textbf{#1}}
\newcommand{\htimesw}[2]{\mbox{$#1$$\times$$#2$}}


\newif\ifshowsolution
\showsolutiontrue

\ifshowsolution  
\newcommand{\Solution}[2]{\paragraph{\bf $\bigstar $ SOLUTION:} {\sf #2} }
\newcommand{\Mistake}[2]{\paragraph{\bf $\blacksquare$ COMMON MISTAKE #1:} {\sf #2} \bigskip}
\else
\newcommand{\Solution}[2]{\vspace{#1}}
\fi

\newcommand{\truefalse}{
\begin{enumerate}
	\item True
	\item False
\end{enumerate}
}

\newcommand{\yesno}{
\begin{enumerate}
	\item Yes
	\item No
\end{enumerate}
}
